\documentclass[authoryear,5p,final,times]{elsarticle}

\pdfoutput=1

\usepackage[utf8]{inputenc}
\usepackage[T1]{fontenc}
\usepackage{amsmath,amsfonts,amssymb}
\usepackage{xfrac,nicefrac}
\usepackage{bm}
\usepackage{csquotes}
\usepackage{nth}
\usepackage{hyperref}
\usepackage{siunitx}
\usepackage{booktabs}
\usepackage{makecell}
\usepackage{adjustbox}
\usepackage{subcaption}
\usepackage{microtype}
\usepackage{ifthen}

\journal{Medical Image Analysis}
\date{January 19, 2019}

\hypersetup{colorlinks=true}
\biboptions{sort&compress}
\DeclareSIUnit{\year}{yr}

\renewcommand{\vec}[1]{\bm{#1}}

\newcounter{fnscnt}
\newcommand{\fns}[1]{%
	\setcounter{fnscnt}{#1}%
	\textsuperscript{\ensuremath{\fnsymbol{fnscnt}}}%
}

\newcommand{\asdmark}{\fns{4}}
\newcommand{\surfacedistance}[2][assd]{%
    \phantom{\asdmark}%
    \num{#2}\,%
    \ifthenelse{\equal{#1}{asd}}{\asdmark}{\phantom{\asdmark}}%
}
\newcommand{\accuracy}[1]{\num{#1}}
\newcommand{\downward}{\raisebox{0.4mm}{\tiny \ensuremath{\downarrow}}}

\begin{document}
    \begin{frontmatter}
        \title{Iterative fully convolutional neural networks for automatic vertebra~segmentation~and~identification}

        \author[isi]{Nikolas Lessmann}
        \author[diag]{Bram van Ginneken}
        \author[rad,uu]{Pim A. de Jong}
        \author[isi]{Ivana I{\v{s}}gum}
                
        \address[isi]{Image Sciences Institute, University Medical Center Utrecht, The~Netherlands}
        \address[diag]{Diagnostic Image Analysis Group, Radboud University Medical Center Nijmegen, The~Netherlands}
        \address[rad]{Department of Radiology, University Medical Center Utrecht, The~Netherlands}
        \address[uu]{Utrecht University, The~Netherlands}
        
        \begin{abstract}
            Precise segmentation and anatomical identification of the vertebrae provides the basis for automatic analysis of the spine, such as detection of vertebral compression fractures or other abnormalities. Most dedicated spine CT and MR scans as well as scans of the chest, abdomen or neck cover only part of the spine. Segmentation and identification should therefore not rely on the visibility of certain vertebrae or a certain number of vertebrae. We propose an iterative instance segmentation approach that uses a fully convolutional neural network to segment and label vertebrae one after the other, independently of the number of visible vertebrae. This instance-by-instance segmentation is enabled by combining the network with a memory component that retains information about already segmented vertebrae. The network iteratively analyzes image patches, using information from both image and memory to search for the next vertebra. To efficiently traverse the image, we include the prior knowledge that the vertebrae are always located next to each other, which is used to follow the vertebral column. The network concurrently performs multiple tasks, which are segmentation of a vertebra, regression of its anatomical label and prediction whether the vertebra is completely visible in the image, which allows to exclude incompletely visible vertebrae from further analyses. The predicted anatomical labels of the individual vertebrae are additionally refined with a maximum likelihood approach, choosing the overall most likely labeling if all detected vertebrae are taken into account. This method was evaluated with five diverse datasets, including multiple modalities (CT and MR), various fields of view and coverages of different sections of the spine, and a particularly challenging set of low-dose chest CT scans. For vertebra segmentation, the average Dice score was \SI{94.9(21)}{\percent} with an average absolute symmetric surface distance of \SI{0.2(1)}{\milli\meter}. The anatomical identification had an accuracy of \SI{93}{\percent}, corresponding to a single case with mislabeled vertebrae. Vertebrae were classified as completely or incompletely visible with an accuracy of \SI{97}{\percent}. The proposed iterative segmentation method compares favorably with state-of-the-art methods and is fast, flexible and generalizable.
        \end{abstract}
    \end{frontmatter}
        
    \section{Introduction}
    
    Segmentation and identification of the vertebrae is often a prerequisite for automatic analysis of the spine, such as detection of vertebral fractures \citep{Yao2012}, assessment of spinal deformities \citep{Forsberg2013}, or computer-assisted surgical interventions \citep{Knez2016}. Automatic spine analysis can be performed with a large variety of tomographic scans, including dedicated spine scans but also scans of the neck, chest or abdomen that incidentally cover part of the spine. A generic vertebra segmentation algorithm therefore needs to be robust with respect to different image resolutions and different coverages of the spine. This especially means that no assumptions should be made about the number of visible vertebrae and their anatomical identity, i.e., to which section of the spine they belong. Vertebra segmentation is therefore essentially an instance segmentation problem with an \emph{a priori} unknown number of instances (\emph{i.e.\ }vertebrae). However, in contrast to generic instance segmentation the individual instances are not independent of each other. The instances are known to be located in close proximity to each other in the image, forming together the vertebral column. We propose to approach vertebra segmentation with an instance segmentation algorithm that explicitly incorporates this prior knowledge to locate instances, but that makes no further assumptions.
    
    Approaching vertebra segmentation as an instance segmentation problem entails treating all vertebrae as instances of the same class of objects. However, an anatomical identification of the segmented vertebrae is often also needed, for instance, for further analysis steps or for reporting purposes. Especially in images originally not intended for spine imaging, anatomical labeling of the vertebrae can be challenging due to variations in the field of view. These variations lead to variable coverage of the spine and also of structures that provide anatomical cues for identification of the vertebrae, such as the ribs or the sacrum. Additionally, neighboring vertebrae often have similar shape and appearance so that independent labeling of each vertebra may result in mistakes. Vertebra identification therefore requires a global rather than a per-instance approach to ensure an overall plausible, anatomically correct labeling.
        
    Another challenge inherent to an instance segmentation approach is the identification of partially visible instances. While occlusion is a typical problem in two-dimensional but not in three-dimensional images, some vertebrae may be only partially visible due to the limited field of view of the scan. If these incompletely visible vertebrae are included in subsequent analyses that are based on the obtained vertebra segmentations, such as measurement of vertebral heights for detection and classification of vertebral compression fractures \citep{Grigoryan2003}, their results may be unreliable. Therefore, incompletely visible instances need to be either ignored or explicitly identified as incomplete so that they can be excluded from subsequent analyses.
    
    In this paper, we propose an iterative instance-by-instance segmentation approach for vertebra segmentation based on a fully convolutional neural network. This network performs vertebra detection, segmentation, anatomical identification and classification of their completeness concurrently and therefore presents an entirely supervised approach that can be trained end-to-end. While we propose to attempt a per-instance identification of the individual vertebrae together with the segmentation, the labeling is subsequently adjusted taking all segmented vertebrae into account. In contrast to previous approaches, the presented method can be used for any imaging modality, any field of view and any number and type (cervical, thoracic, lumbar) of visible vertebrae because it avoids explicit modeling of shape and appearance of the vertebrae and the vertebral column. We evaluate these claims using a diverse selection of datasets, including scans from different modalities (CT and MR), various fields of view, cases with severe compression fractures and a particularly challenging set of low-dose chest CT.
            
    \section{Related work}
        
    While a few other methods have been published that address both vertebra segmentation and identification \citep{Klinder2009,Kelm2013,Chu2015,Suzani2015,Sekuboyina2017}, the majority of methods in the literature focused on one of these problems. The existing literature is therefore reviewed separately for vertebra segmentation and for vertebra identification. We also briefly review literature on general instance segmentation.
    
    \subsection{Vertebra segmentation}
    
    Vertebra segmentation has been approached predominantly as a model-fitting problem using statistical shape models and its variants, most often active shape models and shape-constrained deformable models \citep{CastroMateos2015,Ibragimov2014,Ibragimov2015,Kadoury2011,Kadoury2013,Klinder2009,Korez2015,Korez2016,Mastmeyer2006,Mirzaalian2013,Pereanez2015,Rasoulian2013,Stern2011,Suzani2015,Yang2017,Leventon2002}. Other approaches have been based on atlases \citep{Wang2015}, level-sets with shape priors \citep{Leventon2002,Lim2014} and active contours \citep{Hammernik2015,Athertya2016}.
    
    More recently, machine learning has been increasingly used for vertebra segmentation. \citet{Kelm2013} used an iterative variant of marginal space learning to find bounding boxes for the intervertebral discs, which were used to initialize and guide vertebra segmentations based on Markov random fields and graph cuts. \citet{Zukic2014} applied the Viola-Jones object detection framework based on Adaboost to find bounding boxes for the vertebral bodies, which were subsequently segmented by inflating a mesh from the center of each vertebral body. \citet{Chu2015} used random forest regression to detect the center of the vertebral bodies and used these to define regions of interest in which vertebrae were segmented using random forest voxel classification. A similar method was proposed by \citet{Suzani2015}, who used a multilayer perceptron to regress the distance to the nearest center of a vertebral body. The detected locations were used to initialize an adaptive shape model for segmentation of the vertebral bodies. \citet{Mirzaalian2013} also combined machine learning and shape models by using a probabilistic boosting-tree classifier for boundary detection, which was used to adapt a surface mesh to the vertebrae in combination with a statistical shape model. The shape model was used for initialization of the mesh and to impose shape constraints. \citet{Korez2016} used a convolutional neural network (CNN) to generate probability maps for the vertebral bodies and used these maps to guide a deformable surface model to segment the vertebral bodies.
    
    Even though the aforementioned methods contain a machine learning component beyond statistical modeling, machine learning was primarily used for vertebra detection and thus merely for initialization of the segmentation, which itself was performed with other methods. Many of the most recently published vertebra segmentation methods, however, are based on deep learning and have replaced explicit modeling of the vertebral shape and appearance with convolutional and recurrent neural networks. For instance, \citet{Sekuboyina2017} segmented the lumbar vertebrae in 2D sagittal slices using a multiclass CNN for pixel labeling. As a prior step, a simple multilayer perceptron estimated a bounding box of the lumbar region to identify the region of interest in the image. In subsequent work, \citet{Sekuboyina2018} used a patch-based 3D network for voxel classification in the entire image and additionally a 2D network to predict a low-resolution mask for the vertebral column, which was used to remove false positives outside the spinal region. Similarly, \citet{Janssens2018} relied on two consecutive networks, first using a regression CNN to estimate a bounding box of the lumbar region, followed by a classification CNN to perform voxel labeling within that bounding box to segment the lumbar vertebrae. In our preliminary work \citep{Lessmann2018b}, we also applied a two-stage approach in which vertebrae were first segmented in downsampled images using an iterative strategy. The image was repeatedly analyzed by a CNN to segment the vertebrae one after the other. A second network analyzed the full resolution images to refine the low-resolution segmentations. Even though all of these approaches relied on CNNs for segmentation of the vertebrae, they retained the separation into a detection and a segmentation task, and consequently used two dedicated networks.
        
    \subsection{Vertebra identification}
    
    Anatomical identification of individual vertebrae has been mainly approached with one of three strategies: with appearance and shape models \citep{Klinder2009,Glocker2012,Cai2015}, with machine learning based on hand-crafted features \citep{Chu2015,Glocker2012,Glocker2013,Major2013,Kelm2013,Suzani2015,Bromiley2016} and with deep neural networks \citep{Cai2016,Chen2015,Forsberg2017,Liao2018,Yang2017,Yang2017b,Janssens2018,Sekuboyina2017,Lessmann2018b}. Most of these approaches combined a rough labeling of the vertebrae, typically by performing voxel classification or regression of vertebral centroids or bounding boxes, with a global model to refine the individual predictions, to discard outliers and to find an overall plausible solution. These models have often been graphical models, such as hidden Markov models \citep{Chu2015,Glocker2012} and Markov random fields \citep{Major2013}, statistical shape models \citep{Bromiley2016,Suzani2015,Chen2015} or recurrent neural networks \citep{Yang2017b,Liao2018}. Several methods relied on detection of a reference vertebra, such as the fifth lumbar vertebra (L5), and labeled the detected vertebrae relative to this reference vertebra \citep{Cai2016,Forsberg2017,Lessmann2018b}. Multiclass CNNs for combined segmentation and identification through voxel classification were used in scans with a fixed field of view and a limited number of vertebrae, e.g., lumbar spine CT scans \citep{Sekuboyina2017,Janssens2018}. Furthermore, probabilistic modeling has been used to calculate a likelihood score for each possible labeling configuration based on shape or appearance similarities or spatial relationships of the vertebrae \citep{Glocker2013,Kelm2013,Chen2015,Klinder2009}.
                
    \subsection{Instance segmentation}
    
    Generic instance segmentation frameworks based on convolutional neural networks, such as Mask R-CNN \citep{He2017}, typically split the task into a detection and a segmentation task. Many of the above discussed publications on vertebra segmentation have used this approach as well, but have usually imposed constraints on the number of instances or other features. Recurrent networks have been used in a similar fashion by localizing individual instances based on attention and memory mechanisms, often also based on instance detection through region proposals and subsequent segmentation \citep{Stewart2016,RomeraParedes2016,Li2016,Ren2017}. Other approaches have relied on clustering instead of explicit instance detection, transforming images into abstract feature representations in which individual instances were detected as individual clusters \citep{Uhrig2016,DeBrabandere2017,Liang2018,Novotny2018}. An iterative approach without region proposals or recurrent connections has been proposed by \citet{Li2016}, who repeatedly fed the same image through a convolutional semantic segmentation network together with the previous prediction map.
            
    \section{Methods}
        
    We propose a vertebra segmentation and identification method based on a single fully convolutional neural network (FCN) that performs multiple tasks concurrently. In contrast to existing methods, this avoids a multi-stage process with successive instance detection and segmentation, or segmentation and instance separation steps. Other existing generic instance segmentation methods with 2D deep neural networks often do not generalize well to 3D image volumes because they analyze the entire image at once, which is currently not feasible with typical CT or MR volumes. We therefore apply a patch-based vertebra-by-vertebra segmentation approach in which the image is analyzed in patches large enough to contain at least one vertebra. The network segments a single vertebra in this patch and the anatomical knowledge that the following vertebra must be located in close proximity is used to reposition the patch for segmentation of the following vertebra.
    
\begin{figure*}[t]
        \centering
        \includegraphics[width=0.76\textwidth]{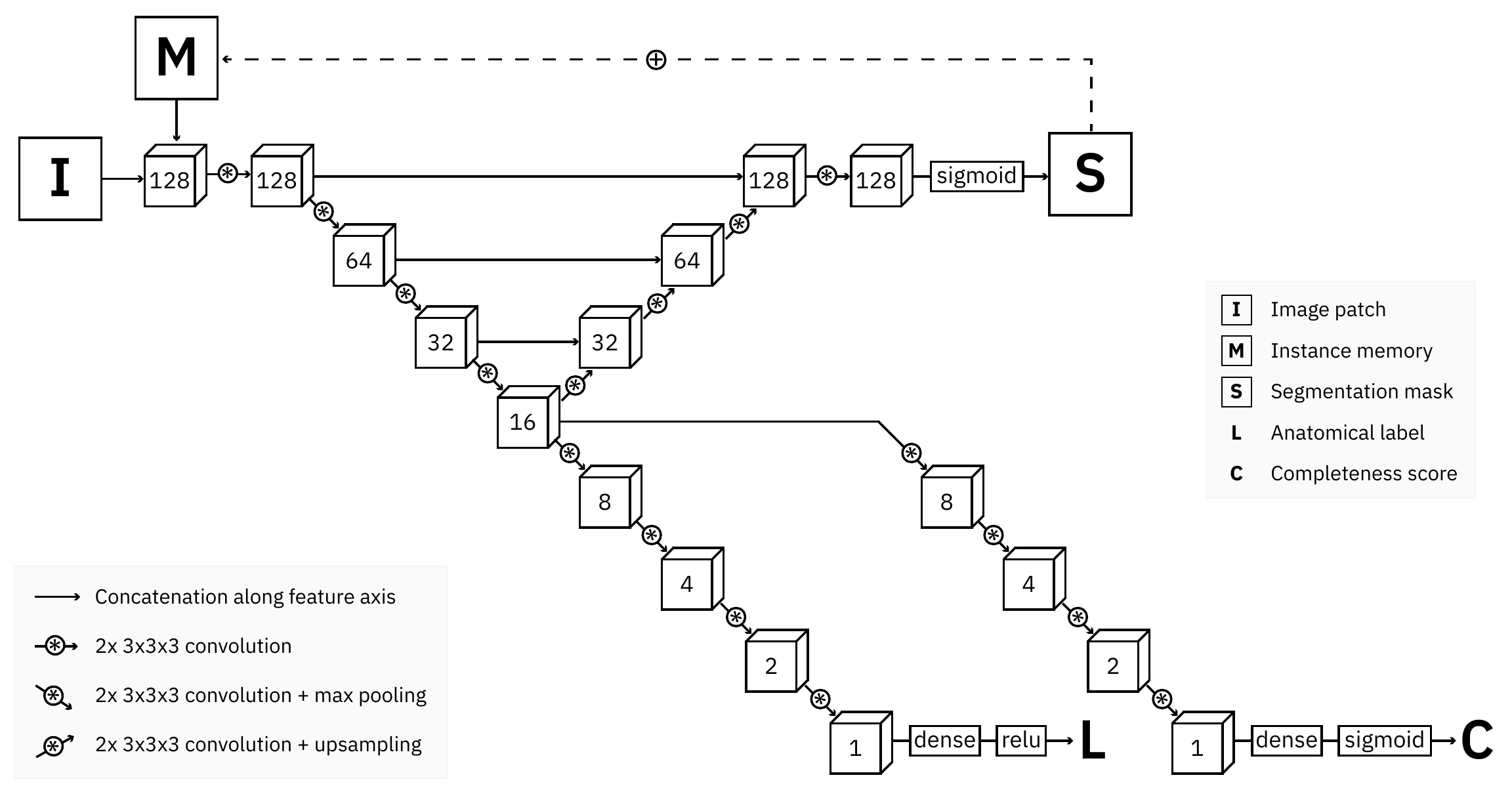}
        \caption{Schematic drawing of the network architecture. I, M and S are 3D volumes, L is the predicted label in form of a single value, and C is the predicted probability for complete visibility. Cubes represent 3D feature maps with 84 channels in the path from I and M to S and 48 features maps in the two additional compression paths to L and C. Exceptions are the first cube after I and M, which has two channels as a result of the concatenation of I and M, and the cube before S, which has only one channel. Both dense layers map 48 features to a single value. The number on each cube indicates the size of the feature map (a cube \enquote{\num{128}} corresponds to a feature map of \num{128x128x128} voxels).}
        \label{fig:network}
    \end{figure*}

    Next to this iterative inference strategy, our approach consists of four major components. The central component is a \emph{segmentation network} that segments voxels from a 3D image patch by binary classification of all voxels in the patch. To enable this network to segment only voxels belonging to a specific instance rather than all vertebrae visible in the patch, we augment the network with an \emph{instance memory} that informs the network about already segmented vertebrae. The network uses this information to always segment only the following not yet segmented vertebra. Once a vertebra is fully segmented, the instance memory is updated, which triggers the network in the next iteration to ignore this vertebra and focus on the next vertebra instead. The third component is an \emph{identification sub-network} that predicts the anatomical label of each detected vertebra. The fourth component is a \emph{completeness classification sub-network} that is added to the network to distinguish between completely visible and partially visible vertebrae. The full network architecture is illustrated in Figure~\ref*{fig:network}. Please note that the aforementioned components, referred to as networks and sub-networks, together form a single network.
  
    In the following, we will first describe the segmentation network (section~\ref{sec:method_segmentation}) and the instance memory (section~\ref{sec:method_memory}). These two components form the basis for the proposed iterative instance segmentation strategy (section~\ref{sec:method_iterative}). Additional network parts without influence on the iterative segmentation are described afterwards: regression of the anatomical label of each vertebra (section~\ref{sec:method_identification}) and classification of its completeness (section~\ref{sec:method_completeness}). Finally, the loss function to optimize the network for all three tasks and the training process are detailed (section~\ref{sec:method_training}).
    
    \subsection{Segmentation network}
    \label{sec:method_segmentation}
    
    The segmentation component of the network is a FCN that predicts a binary label for each voxel in an image patch. This label indicates whether the voxel belongs to the current instance or not. We used a patch size of \num{128x128x128} voxels, which is large enough to cover an entire vertebra. To ensure that all patches have the same resolution, even though the analyzed images may have different resolutions, we resample all input images to an isotropic resolution of \SI{1x1x1}{\milli\meter} prior to the segmentation. The obtained segmentation masks are resampled back to the original image resolution using nearest neighbor interpolation.
    
    The architecture of the segmentation network is inspired by the U-net architecture \citep{Ronneberger2015,Cicek2016}, i.e., the network consists of a compression and an expansion path with intermediate skip connections. We use a constant number of filters in all layers and added batch normalization as well as additional padding before all convolutional layers to obtain segmentation masks of the same size as the image patches. The segmentation masks are additionally refined in CT scans by removing voxels below \SI{200}{HU} from the surface of each vertebra.
    
    \subsection{Instance memory}
    \label{sec:method_memory}
    
    In the proposed iterative segmentation scheme, which is described in detail in the following section, the network segments one vertebra after the other, one at a time. The purpose of the instance memory is to remind the network of vertebrae that were already segmented in previous iterations so that it can target the next not yet segmented vertebra. This memory is a binary flag for each voxel of the input image that indicates whether the voxel has been labeled as vertebra in any of the previous iterations. Binary rather than probabilistic flags are used for simplicity, but probabilistic flags could prove useful in future extensions that allow voxels labeled as part of a vertebra to be unlabeled or relabeled in later iterations. Together with an image patch, the network receives a corresponding memory patch as input. These are fed to the network as a two-channel input volume.
    
    \subsection{Iterative instance segmentation}
    \label{sec:method_iterative}
    
\begin{figure*}[t]
        \centering
        \includegraphics[width=0.76\textwidth]{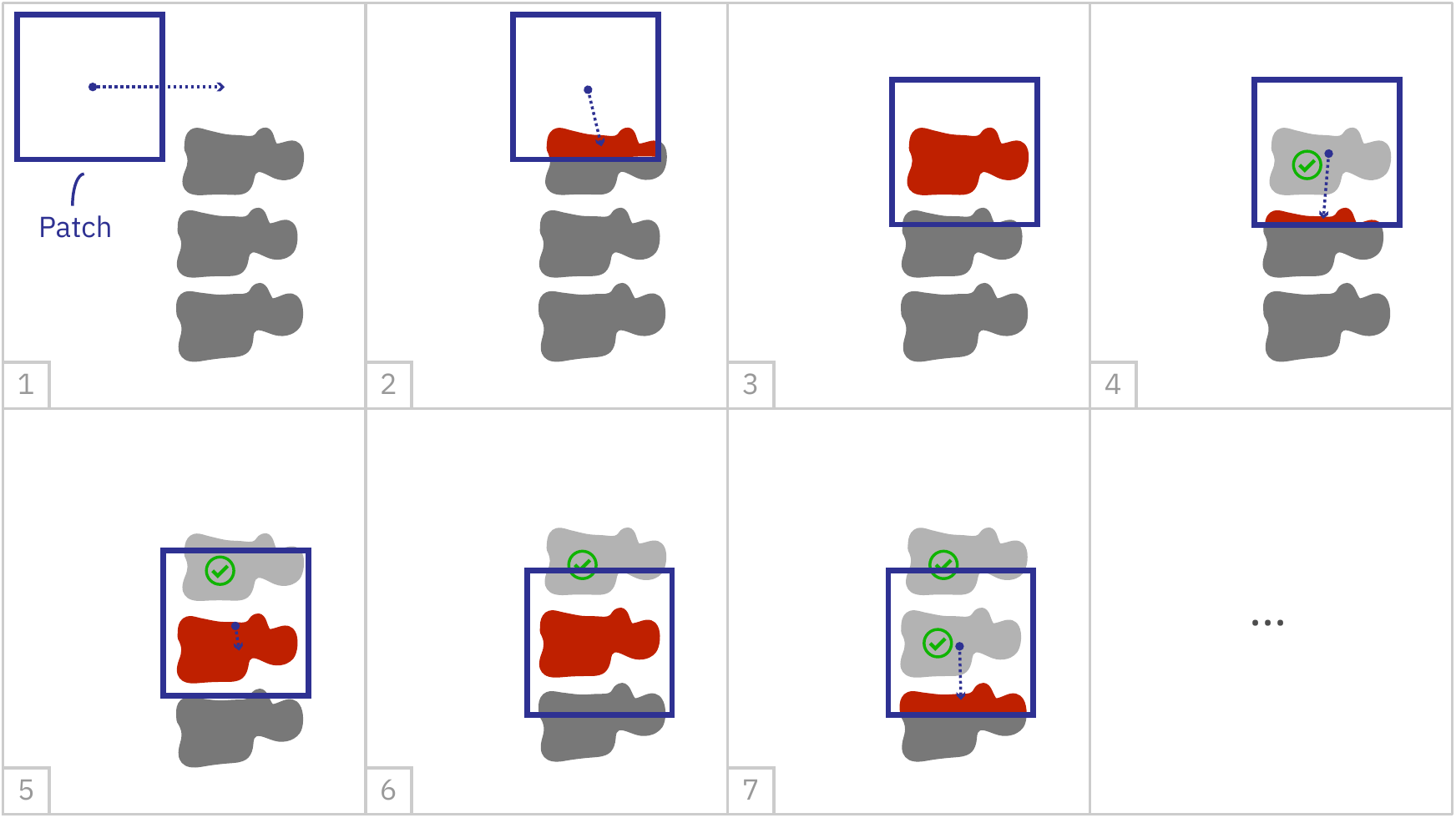}
        \caption{Illustration of the iterative instance segmentation and traversal strategy. The patch is first moving in a sliding window fashion over the image (1), until a fragment of vertebral bone is detected (2). The patch is then moved to the center of the detected fragment. This process is repeated until the entire vertebra becomes visible and the patch thus stops moving (3). The segmented vertebra is added to the instance memory and the same patch is analyzed again, now yielding a fragment of the following vertebra because the updated memory forces the network to ignore the previous vertebra (4). The patch is centered now at the detected fragment of the following vertebra and the process repeats (5-7).}
        \label{fig:itseg}
    \end{figure*}

    The iterative segmentation process is illustrated in Figure~\ref*{fig:itseg}. This process follows either a top-down or bottom-up scheme, i.e., the vertebrae are not segmented in random order but successively from top to bottom, or vice versa. The network learns to infer from the memory patch which vertebra to segment in the current patch. If the memory is empty, i.e., no vertebra has been detected yet, the network segments the top-most or bottom-most vertebra that is visible in the patch, depending on the chosen direction of traversal. Otherwise, the network segments the first following not yet segmented vertebra, even if multiple unsegmented vertebrae are visible. Other vertebrae that are the second or third not yet segmented vertebra in the direction of traversal are disregarded until they become the first not yet segmented vertebra themselves in a later iteration. Each instance of the network therefore has a fixed direction of traversal.
    
    The patch size is chosen large enough to always contain part of the following vertebra when a vertebra is in the center of the patch. This enables utilizing prior knowledge about the spatial arrangement of the individual instances to move from vertebra to vertebra. The FCN iteratively analyzes a single patch centered at $\vec{x}_t$, where $t$ denotes the iteration step. Initially, the patch is moved over the image in a sliding window fashion with constant step size $\Delta \vec{x}$, searching for the top-most vertebra when using a top-down approach, or the bottom-most vertebra when using a bottom-up approach. As soon as the network detects a large enough fragment of vertebral bone, at least $v_\text{min} = 1000\,\text{voxels} \mathrel{\hat{=}} 10\,\text{mm}^3$ in our experiments, the patch is moved toward this fragment. The center of the bounding box of the detected fragment, referred to as $\vec{b}_t$, becomes the center of the next patch:
    
    \begin{equation*}
        \vec{x}_{t+1} =
        \begin{cases}
          \vec{x}_t + \Delta \vec{x}, & \text{if}\ v_t < v_{\text{min}}\\
          [\, \vec{b}_t \,], & \text{otherwise}\\
        \end{cases}
    \end{equation*}
    
    Even if initially only a small part of the vertebra is visible and detected in the patch, centering the following patch at the detected fragment ensures that a larger part of the vertebra becomes visible in the next iteration. Eventually, the entire vertebra becomes visible, in which case the patch position converges because no additional voxels are detected anymore that would affect $\vec{b}_t$, and hence $\vec{x}_{t+1}$. We detect convergence by comparing the current patch position $\vec{x}_t$ and the previous patch position $\vec{x}_{t-1}$, testing whether they still differ by more than $\delta_\text{max}$ on any axis. Occasionally, the patch position does not converge but keeps alternating between positions that differ slightly more than the threshold $\delta_\text{max}$. We therefore limit the number of iterations per vertebra. When this limit is reached, we assume that the patch has converged to the position between the two previous patch positions and we accordingly move the patch to $
        \vec{x}_{t+1} =
        \left[ \ \left(
          \vec{x}_{t} + \vec{x}_{t-1}
        \right) /\ 2 \ \right]
    $. In our experiments, $\delta_\text{max}$ was set to \num{2}, $\Delta \vec{x}$ to half the patch size and the maximum number of iterations to \num{10}.
    
    Once the position has converged, the segmented vertebra is added to the output mask using a unique instance label and the segmentation mask obtained at this final position. Furthermore, the instance memory is updated. In the following iteration, the network analyzes the same patch again. The updated memory prompts the network to detect a fragment of the following vertebra and the patch is moved to the center of the detected new fragment, repeating the segmentation process for the next vertebra. Should no fragment of the following vertebra be immediately visible, traversal reverts to a sliding window motion until the next vertebra is found. The entire process continues until no further fragments are found, i.e., until all visible vertebrae are segmented.
    
    \subsection{Anatomical identification}
    \label{sec:method_identification}
    
    For each detected vertebra, the network additionally predicts the anatomical label using an additional identification component that is added to the segmentation network. The U-net architecture, which provides the basis of the segmentation network, consists of a compression and a expansion path. These are commonly understood as recognition and segmentation paths, respectively. Since vertebra identification is a recognition task, the identification component is appended to the compression path as further compression steps. Essentially, it shares features with the segmentation network but further compresses the input patch into a single value. The vertebrae C1 to L5 are represented by integers \numrange{1}{24} and the value \num{0} is used during training as ground truth value for patches that do not contain any vertebral bone. In the network, the label output unit is a rectified linear unit $f(x) = \max \left( 0, x \right)$, which enables the network to produce any negative activation for patches without vertebral bone. Regression of the label as continuous value instead of classification with \num{25} output units and a softmax activation function ensures that the loss function penalizes predictions more strongly the further they deviate from the true label.
    
    In order to find a plausible sequence of labels for the detected vertebrae, without duplicates or gaps, the predicted labels are interpreted as probabilities in a simple maximum likelihood estimation similar to \citet{Klinder2009}. Depending on how many vertebrae are detected, several labeling sequences are theoretically possible -- only if all \num{24} vertebrae are visible in a scan there is only a single possible labeling sequence. For each possible sequence of labels, the average likelihood is calculated by interpreting the predicted labels as probabilities. For example, a regression output value of \num{22.8} is interpreted as \SI{80}{\percent} probability for label \num{23} (L4) and \SI{20}{\percent} probability for label \num{22} (L3), and as \SI{0}{\percent} probability for any other label. In sequences in which this vertebra would be labeled as L3, it therefore contributes a probability of \SI{20}{\percent} to the average. The sequence with maximal likelihood is finally used to labeled the detected vertebrae. This step is performed as a post-processing step when all vertebrae have been segmented from the image.
    
    \subsection{Classification of vertebra completeness}
    \label{sec:method_completeness}
    
    An obvious strategy for disregarding incompletely visible vertebrae in the segmentation process would be to train the network only with examples of fully visible vertebrae. However, the traversal scheme requires detection of vertebral fragments in the patches. We therefore choose to include partially visible vertebrae in the training data but to add a classification component to the segmentation network that classifies each segmented vertebra as complete or incomplete. Similar to the identification component, this completeness classification is a recognition and not a segmentation task. In the network, the classification path is therefore also a continuation of the compression path and has the same architecture as the identification path, with the difference that the output is a sigmoid unit. The output value is thus a single value in $[0,1]$, which indicates the probability that the vertebra is completely visible in the image patch. During traversal, vertebrae classified as incomplete are not added to the segmentation mask, but are added to the instance memory to facilitate further traversal because not only the last but also the first vertebrae are often incompletely visible. This additionally ensures that traversal can continue if a vertebra is mistakenly classified as incompletely visible. Depending on the application, these vertebrae could also be added to the segmentation mask, merely flagging them as incomplete but making them available to further analysis steps.
    
    \subsection{Training the network}
    \label{sec:method_training}
    
    During training, we derived the status of the instance memory from the reference segmentation masks. The patches used to train the network were forced to contain vertebral bone by randomly selecting in each iteration a scan and a vertebra visible in that scan, followed by random patch sampling within the bounding box of that vertebra. However, \SI{25}{\percent} of the patches were selected randomly from anywhere in the training images. If these patches contained vertebral bone, it was added to the instance memory so that the network could also learn to produce empty segmentation masks for patches without vertebral bone or without unsegmented vertebral bone.
    
    Due to the size of the input patches, the Nvidia Titan X GPUs with 12 GB memory that we used for training allowed processing of only single patches instead of minibatches of multiple patches. We therefore used Adam \citep{Kingma2014} for optimization with a fixed learning rate of \num{0.001} and an increased momentum of \num{0.99}, which stabilizes the gradients. Furthermore, the network predicts labels for all voxels in the input patch and the loss term is accordingly not based on a single output value, but on more than two million output values per patch. The loss term $\mathcal{L}$ combines terms for segmentation, anatomical labeling and classification errors:
    
    \begin{equation*}
        \begin{aligned}
            \mathcal{L} & =
            \underbrace{\strut \:\lambda \cdot \textnormal{FP}_{\textnormal{soft}} + \textnormal{FN}_{\textnormal{soft}}\:}_\text{Segmentation error}
            \:+\:
            \underbrace{\strut \:|\, p_\textnormal{\tiny L} - t_\textnormal{\tiny L}\, |\:}_\text{Labeling error}
            \\ &
            \:+\:
            \underbrace{\strut \left(-t_\textnormal{\tiny C} \, \log p_\textnormal{\tiny C} - (1 - t_\textnormal{\tiny C}) \log (1 - p_\textnormal{\tiny C})\right)\:}_\text{Completeness classification error}
        \end{aligned}
    \end{equation*}
    
    We propose to optimize the segmentation by directly minimizing the number of incorrectly labeled voxels, i.e., the number of false positives and false negatives. This is similar to loss terms based on the Dice score \citep{Milletari2016}, which can be expressed as $\sfrac{2\,\textnormal{TP}}{\left(2\,\textnormal{TP} + \textnormal{FP} + \textnormal{FN}\right)}$, but because the number of true positives is not part of the term values are more consistent across empty ($\textnormal{TP} = 0$) and non-empty ($\textnormal{TP} \gg 0$) patches. Given an input patch $I$ and for all voxels $i$ binary reference labels $t_i$ and probabilistic predictions $p_i$, differentiable expressions for the number of false positive and false negative predictions are:
    
    \begin{equation*}
        \textnormal{FP}_{\textnormal{soft}} = \sum_{i \in I} \omega_{i} \cdot (1 - t_{i}) \, p_{i}
        \quad\enspace
        \textnormal{FN}_{\textnormal{soft}} = \sum_{i \in I} \omega_{i} \cdot t_{i} \, (1 - p_{i})
    \end{equation*}
    
    \noindent
    Here, $w_i$ are weights used to assign more importance to the voxels near the surface of the vertebra. This aims at improving the separation of neighboring vertebrae \citep{Ronneberger2015}. The weights are derived from the distance $d_i$ of voxel $i$ to the closest point on the surface of the targeted vertebra: $
        \omega_i = \gamma \cdot \exp \left( -\, \nicefrac{d_i^2}{\sigma^2} \right) + 1
    $. We used $\gamma = 8$ and $\sigma = 6$ in all experiments.
    
    Additionally, the factor $\lambda$ weights the cost of a false positive error relative to a false negative error, which is useful to counteract an imbalance between background and foreground voxels. In most segmentation problems, the number of background voxels is substantially larger than the number of foreground voxels and consequently a systematic false negative mistake is more favorable than a systematic false positive mistake. This can prevent the network from learning anything other than predicting an empty segmentation mask for any input. We therefore used an increasing value for $\lambda$ starting from $\lambda_{\textnormal{min}} = 0.1$ and converging eventually at 1, following a sigmoidal shape:
    
    \begin{equation*}
        \lambda(n) =
            \lambda_{\textnormal{min}} +
            \frac{
                1 - \lambda_{\textnormal{min}}
            }{
                1 + e^{-\vartheta(n)}
            }
            \quad
            \textnormal{where}
            \quad
            \vartheta(n) = \frac{
                n - \sfrac{n_{\textnormal{max}}}{2}
            }{
                \sfrac{n_{\textnormal{max}}}{10}
            }
    \end{equation*}
    
    \noindent
    Here, $n$ is the number of training iterations, i.e., backward passes, that have been performed. The definition of $\vartheta$ ensures that false positive and false negative errors are equally weighted after about half of the maximum number of training iterations $n_{\textnormal{max}}$. The network is therefore initially biased towards making false positive errors rather than false negative errors, but this bias is reduced over time to ensure that the network will not converge in a state where it tends to oversegment the vertebra. We found that assigning weights to the individual error components that form the loss $\mathcal{L}$ was not necessary.
    
    The labeling error is defined as the $\ell 1$ norm of the difference between predicted label and true label. The predicted labels are real numbers $\geq 0$ while the true labels are integers \numrange{0}{24}. Using the absolute difference of these two values as loss function penalizes large errors more strongly than small errors. This is especially beneficial because small deviations can be more likely corrected in the label refinement step.
    
    The completeness classification error is defined as the binary cross entropy between the true label $t_\textnormal{\tiny C}$, which is a binary value that indicates whether the vertebra is completely visible in the patch, and the predicted probability for complete visibility $p_\textnormal{\tiny C}$. In our experiments, vertebrae were considered completely visible in a patch when they were manually marked as completely visible in the scan and when not more than \SI{2}{\percent} of the volume of the reference segmentation of the vertebra was not contained in the patch. This allows for some tolerance as manual identification of incompletely visible vertebrae can be ambiguous in scans with low resolution or low-dose artifacts.
    
    We used random elastic deformations, random Gaussian noise, random Gaussian smoothing as well as random cropping along the z-axis to augment the training data. Rectified linear units were used as activation function in all layers except the output layers of the segmentation and the classification paths, in which sigmoid functions were used. The network was implemented using the PyTorch framework. Training on an Nvidia Titan X GPU took about 4--5 days when training for \num{100000} iterations.
        
    \section{Evaluation}
    
    \subsection{Datasets}
    
\begin{figure}[t]
        \centering
        \resizebox{\columnwidth}{!}{
	        \begin{subfigure}[b]{2.4cm}
	            \centering
	            \includegraphics[height=8.1cm]{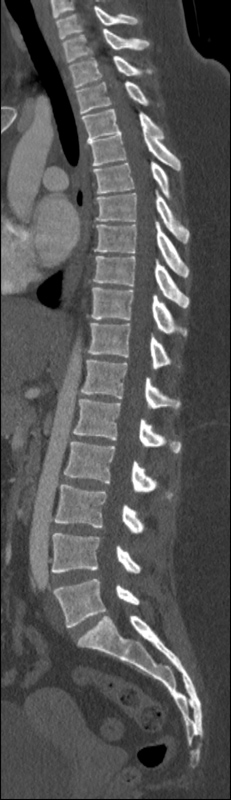}
	            \caption{}
	        \end{subfigure}
	        \hspace{1.15em}
	        \begin{subfigure}[b]{4.28cm}
	        	\begin{minipage}[b][8.1cm]{\textwidth}
		            \centering
		            \includegraphics[height=3.6cm]{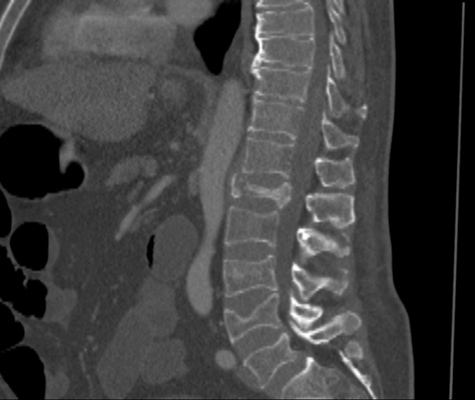}
		            \caption{}
		            
		            \vspace*{\fill}
		            
		            \includegraphics[height=3.6cm]{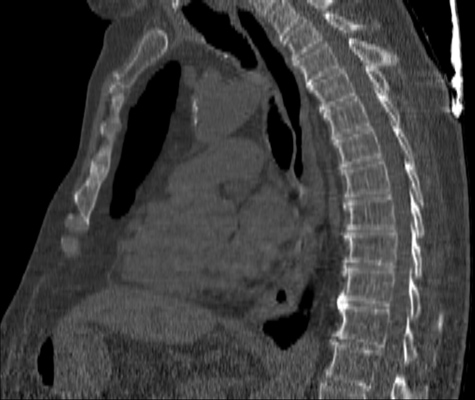}
	        	\end{minipage}
	            	\caption{}
	        \end{subfigure}
	        \hspace{1.15em}
	        \begin{subfigure}[b]{3.4cm}
	        	\begin{minipage}[b][8.1cm]{\textwidth}
		            \centering
		            \includegraphics[height=3.6cm]{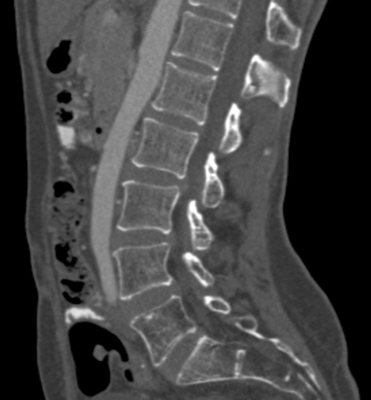}
		            \caption{}
		            
		            \vspace*{\fill}
		            
		            \includegraphics[height=3.6cm]{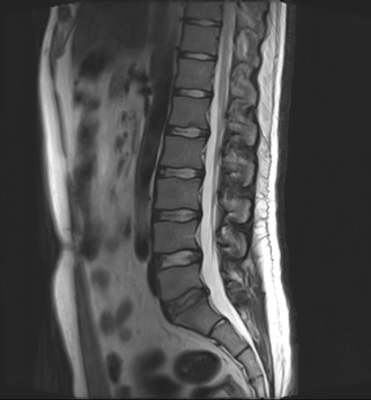}
			    \end{minipage}
	            	\caption{}
	        \end{subfigure}
        }
        \caption{Examples of the various image types. Shown are examples of (a) thoracolumbar spine CT, (b) lumbar spine CT with multiple compression fractures, (c) low-dose chest CT, (d) lumbar spine CT, and (e) lumbar spine MR (T2-weighted).}
        \label{fig:examples_plain_images}
    \end{figure}

    We trained and evaluated the method with five sets of CT and MR scans that visualize the spine. Reference segmentation masks for four of these datasets are publicly available, which allowed for a comparison with other publications that used the same data. Examples of images from the datasets are shown in Figure~\ref*{fig:examples_plain_images}.
    
    The \emph{thoracolumbar spine CT} dataset consists of 15 dedicated spine CT scans that visualize all thoracic and lumbar vertebrae. It was originally used for the spine segmentation challenge held in conjunction with the Computational Spine Imaging (CSI) workshop at MICCAI 2014 \citep{Yao2016}. All subjects were young adults (\numrange{20}{34} years) without vertebral fractures who were scanned with IV-contrast administration. The scans were reconstructed to in-plane resolutions of \SIrange{0.31}{0.36}{\milli\meter} and slice thicknesses of \SIrange{0.7}{1.0}{\milli\meter}. Semi-automatically obtained reference segmentations were provided by the challenge organizers. To allow for a comparison with the challenge results, we used the same data split with 5 scans for evaluation and the remaining 10 scans for training and development.
        
    The \emph{xVertSeg.v1} dataset consists of 15 lumbar spine CT scans of subjects with compression fractures of various grades and types \citep{Ibragimov2017}. Manual reference segmentations are available for the lumbar vertebrae and were defined through a consensus reading of two observers. The scans were reconstructed to in-plane resolutions of \SIrange{0.29}{0.80}{\milli\meter} and slice thicknesses of \SIrange{1.0}{1.9}{\milli\meter}. There are currently two other publications that used the same dataset, but with different evaluation/training separation \citep{Janssens2018,Sekuboyina2017}. We therefore used the scans \numrange{1}{5} for evaluation and the remaining 10 scans for training.
    
    The \emph{low-dose chest CT} dataset consists of 55 scans from the National Lung Screening Trial \citep{NLST2011}. These scans were acquired for lung imaging and visualize in addition to the lungs a variable section of the thoracic and upper lumbar vertebrae. The scanned subjects were heavy smokers aged 50 to 74 years and therefore at increased risk for vertebral compression fractures due to their advanced age and smoking history. The scans were acquired with low radiation dose and reconstructed to in-plane resolutions of \SIrange{0.54}{0.82}{\milli\meter} and slice thicknesses of \SIrange{1.0}{2.5}{\milli\meter}. We created manual and semi-automatic reference segmentations for this dataset: 10 scans were used for evaluation and were therefore fully manually annotated by drawing along the contour of each vertebra in sagittal slices using an interactive live wire tool \citep{Barrett1997}. The contours were converted into segmentation masks, in which inaccuracies and other mistakes were corrected voxel-by-voxel. An additional set of 5 scans was annotated in the same way and was used to train a preliminary version of the network. This network was used to predict rough segmentations in the remaining 40 scans. These rough segmentations were manually inspected and corrected voxel-by-voxel, and were used for training of the final network. This strategy enabled us to create a large training set with substantially less manual annotation effort compared to fully manual segmentation, which is not necessarily needed for training data. Additionally, a second observer fully manually annotated two scans from the evaluation set for an estimation of the interobserver agreement. All fully manual and semi-automatic segmentations were performed in sagittal views by observers who received detailed instructions beforehand. Additionally, all segmentations were validated by an experienced radiologist.
    
    The \emph{lumbar spine CT} dataset consists of 10 scans of healthy subjects and corresponding manual reference segmentations of the lumbar vertebrae \citep{Ibragimov2014,Korez2015}. The scans were reconstructed to in-plane resolutions of \SIrange{0.28}{0.79}{\milli\meter} and slice thicknesses of \SIrange{0.7}{1.5}{\milli\meter}. Because this dataset is the smallest of the datasets that we included, it was used for an external evaluation of our supervised approach. Scans from this dataset were therefore only used for evaluation and were not part of the training set.
    
    The \emph{lumbar spine MR} dataset consists of 23 T2-weighted turbo spine echo MR images acquired at 1.5T in sagittal orientation \citep{Chu2015}. The scans have a resolution of \SI[allow-number-unit-breaks]{2x1.25x1.25}{\milli\meter}. Manual reference segmentations are available for 7 vertebrae (T11-L5) in all scans. These reference segmentations contain only the vertebral bodies, not the entire vertebrae.
    
    Some scans in these datasets contained vertebrae or fragments of vertebrae that were not included in the reference segmentation, e.g., scans for which segmentations of only the lumbar vertebrae were available but in which also some thoracic vertebrae were visible. We manually cropped these images to restrict the field of view to the vertebrae that were included in the reference segmentation in order to avoid training the network with vertebra voxels incorrectly labeled as background voxels. In all scans, the segmented vertebrae were also anatomically labeled and marked as either completely or incompletely contained in the scan. None of the datasets contained multiple scans of the same subject.
    
    \subsection{Evaluation metrics}
    
    The segmentation performance was evaluated with the metrics most commonly reported in publications that used the same datasets. These metrics were the Dice coefficient to measure volume overlap and the average absolute symmetric surface distance (ASSD) to measure segmentation accuracy along the vertebral surface. Both metrics were calculated for individual vertebrae and then averaged across all scans. Each vertebra in the reference segmentation was compared with the vertebra in the automatic segmentation mask with which it had the largest overlap.
    
    The identification performance was evaluated using the identification accuracy, i.e., the percentage of vertebrae that were assigned the correct anatomical label, and the linearly weighted kappa coefficient \citep{Cohen1968}. Through the linear weighting, the kappa coefficient captures also the magnitude of mistakes.
    
    The completeness classification performance was evaluated using the classification accuracy and the average number of false positives and false negatives per scan. False positives were in this case vertebrae that were incompletely visible, but were classified as completely visible.
    
    \section{Experiments and Results}
        
    We trained modality-specific instances of the network adjusted to the different ground truths, i.e., to perform vertebra segmentation in CT and vertebral body segmentation in MR. The CT training set consisted of 60 scans, of which 10 were thoracolumbar spine CT, 10 lumbar spine CT scans with compression fractures and 40 NLST scans. The CT evaluation set consisted of 30 scans, of which 5 were thoracolumbar spine CT scans, 5 lumbar spine CT with compression fractures, 10 NLST scans, and 10 normal lumbar spine CT scans, which were not represented in the training set. The dataset for vertebral body segmentation in MR consisted of only 23 scans in total and we therefore performed 3-fold cross-validation with evaluation sets of 8, 8 and 7 scans and the remaining scans used for training. The MR images were normalized by clipping off values below the \nth{5} and above the \nth{95} percentile and transforming the values into the range $[-1,1]$. We compared the performance in vertebra segmentation, identification and completeness classification with that of other methods in the literature that were evaluated on the same datasets (Table~\ref*{tbl:results}). The networks were trained to traverse the spine upwards.
    
    \begin{table*}[t]
    	\caption{Quantitative results of automatic segmentation, anatomical identification and completeness classification. Data are reported as mean $\pm$ standard deviation and were obtained with modality-specific networks, i.e., with a network trained with CT images and another network trained with MR images.}
    	\centering
    	\resizebox{0.84\linewidth}{!}{
    		\begin{tabular}{lccccccc}
    			\toprule
    			& \multicolumn{2}{c}{\emph{Segmentation}} & \multicolumn{2}{c}{\emph{Identification}} & \multicolumn{3}{c}{\emph{Completeness classification}} \\
    			\addlinespace[0.2em]
    			Dataset &
    			Dice score (\si{\percent}) &
    			ASSD (\si{\milli\meter}) &
    			Accuracy (\si{\percent}) &
    			$\kappa$ &
    			Accuracy (\si{\percent}) &
    			FP\,/\,scan &
    			FN\,/\,scan \\
    			\midrule
    			Thoracolumbar spine CT &
    			&   
    			&   
    			&   
    			&   
    			&   
    			&   
    			\\  
    			\hspace{0.25cm} Proposed method &
    			\num{96.3(13)} &
    			\surfacedistance{0.1(1)} &
    			\accuracy{81} & 
    			\num{0.97} &
    			\accuracy{100} &
    			\num{0} &
    			\num{0} \\
    			\hspace{0.25cm} \citet{Lessmann2018b} & 
    			\num{94.8(16)} &
    			\surfacedistance{0.3(1)} &
    			- &
    			- &
    			- &
    			- &
    			- \\
    			\hspace{0.25cm} \citet{Korez2015} & 
    			\num{94.4(21)} &
    			\surfacedistance{0.3(1)} &
    			- &
    			- &
    			- &
    			- &
    			- \\
    			\addlinespace[0.5em]
    			Lumbar spine CT\,\fns{1} & & & & & & & \\
    			\hspace{0.25cm} Proposed method &
    			\num{94.6(22)} & 
    			\surfacedistance{0.3(2)} &
    			\accuracy{100} & 
    			\num{1.0} &
    			\accuracy{96.0} & 
    			\num{0} & 
    			\num{0.2} \\
    			\hspace{0.25cm} \citet{Janssens2018} &
    			\num{95.7(8)} &
    			\surfacedistance{0.4(1)} &
    			- &
    			- &
    			- &
    			- &
    			- \\
    			\hspace{0.25cm} \citet{Sekuboyina2017} &
    			\num{94.3(28)} &
    			- &
    			- &
    			- &
    			- &
    			- &
    			- \\
    			\addlinespace[0.5em]
    			Low-dose chest CT & & & & & & & \\
    			\hspace{0.25cm} Proposed method &
    			\num{93.1(15)} & 
    			\surfacedistance{0.3(1)} &
    			\accuracy{100} & 
    			\num{1.0} &
    			\accuracy{95.7} & 
    			\num{0.5} & 
    			\num{0.1} \\
    			\hspace{0.25cm} Second observer\,\fns{2} &
    			\num{96.4(6)} &
    			\surfacedistance{0.2(0)} &
    			\accuracy{100} & 
    			\num{1.0} &
    			\accuracy{100} &
    			\num{0.0} &
    			\num{0.0} \\
    			\addlinespace[0.5em]
    			Lumbar spine CT & & & & & & & \\
    			\hspace{0.25cm} Proposed method &
    			\num{96.5(8)} & 
    			\surfacedistance{0.2(0)} &
    			\accuracy{100} & 
    			\num{1.0} &
    			\accuracy{100} & 
    			\num{0.0} & 
    			\num{0.0} \\
    			\hspace{0.25cm} \citet{Korez2015} &
    			\num{95.3(14)} &
    			\surfacedistance{0.3(1)} &
    			- &
    			- &
    			- &
    			- &
    			- \\
    			\hspace{0.25cm} \citet{Chu2015} &
    			\num{91.0(70)} &
    			\surfacedistance[asd]{0.9(3)} &
    			- &
    			- &
    			- &
    			- &
    			- \\
    			\hspace{0.25cm} \citet{Ibragimov2014} &
    			\num{93.6(11)} &
    			\surfacedistance{0.8(1)} &
    			- &
    			- &
    			- &
    			- &
    			- \\
    			\addlinespace[0.3em]
    			\midrule
    			\addlinespace[0.5em]
    			Lumbar spine MR\,\fns{3} & & & & & & & \\
    			\hspace{0.25cm} Proposed method &
    			\num{94.4(33)} & 
    			\surfacedistance{0.4(3)} &
    			\accuracy{100} & 
    			\num{1.0} &
    			\accuracy{100} & 
    			\num{0} & 
    			\num{0} \\
    			\hspace{0.25cm} \citet{Korez2016} &
    			\num{93.4(17)} &
    			\surfacedistance{0.5(1)} &
    			- &
    			- &
    			- &
    			- &
    			- \\
    			\hspace{0.25cm} \citet{Chu2015} &
    			\num{88.7(29)} &
    			\surfacedistance[asd]{1.5(2)} &
    			- &
    			- &
    			- &
    			- &
    			- \\
    			\bottomrule
    			\addlinespace[0.5em]
    			\multicolumn{8}{r}{%
    				\footnotesize
    				\fns{1}\,xVertSeg.v1 dataset
    				\enspace
    				\fns{2}\,subset (2/10 scans)
    				\enspace
    				\fns{3}\,only vertebral bodies
    				\enspace
    				\fns{4}\,ASD (non-symmetric)
    			} \\
    		\end{tabular}
    	}
    	\label{tbl:results}
    \end{table*}
    
	\subsection{Segmentation performance}
	
\begin{figure*}
		\centering
		\scriptsize
		\includegraphics[width=0.242\textwidth]{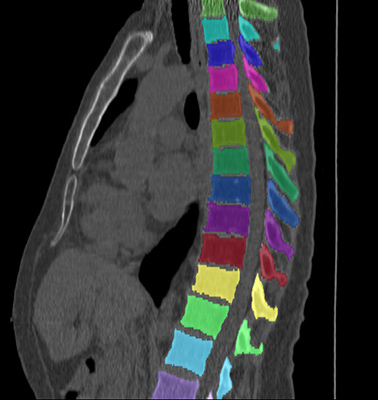}%
		\,%
		\includegraphics[width=0.242\textwidth]{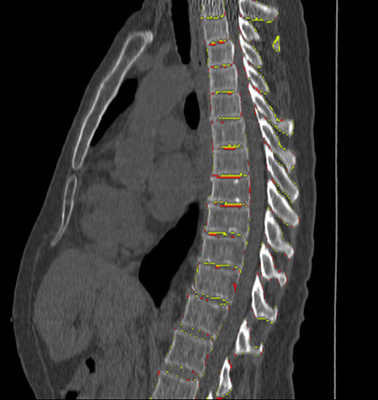}%
		\hfill%
		\includegraphics[width=0.242\textwidth]{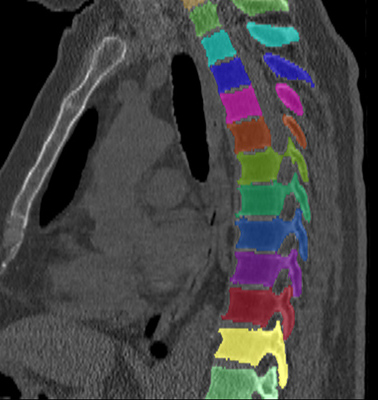}%
		\,%
		\includegraphics[width=0.242\textwidth]{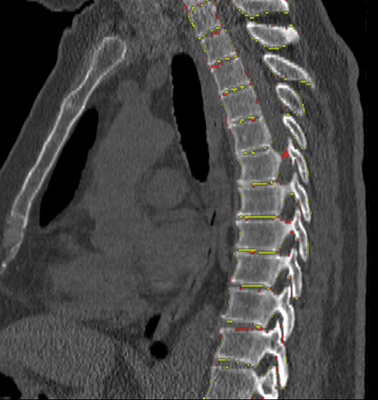}%
		\\[0.25em]
		(a) Low-dose chest CT
		\\[1em]
		\includegraphics[width=0.242\textwidth]{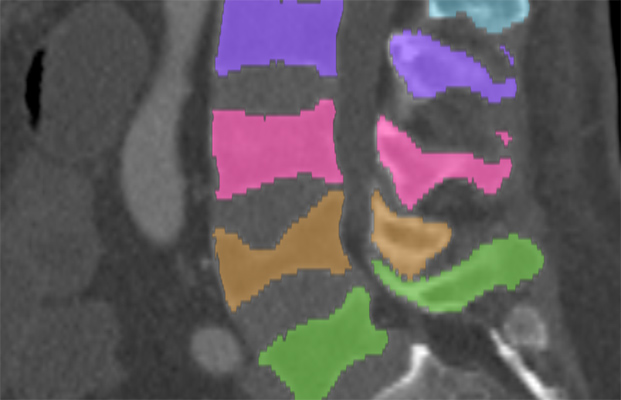}%
		\,%
		\includegraphics[width=0.242\textwidth]{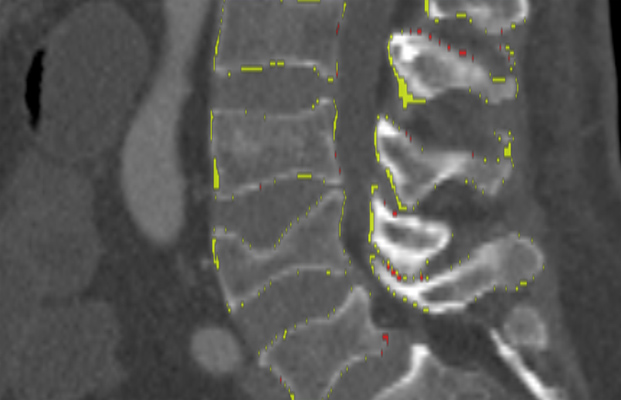}%
		\hfill%
		\includegraphics[width=0.242\textwidth]{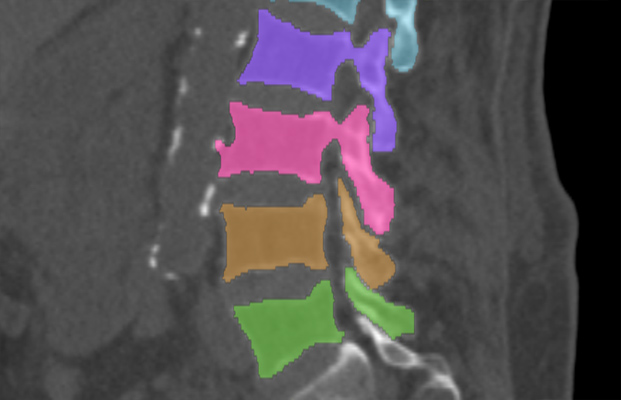}%
		\,%
		\includegraphics[width=0.242\textwidth]{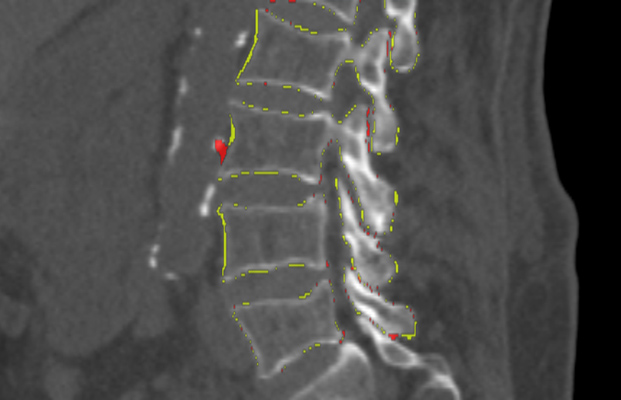}%
		\\[0.25em]
		(b) Lumbar spine CT (xVertSeg.v1 dataset)
		\\[1em]%
		\includegraphics[width=0.242\textwidth]{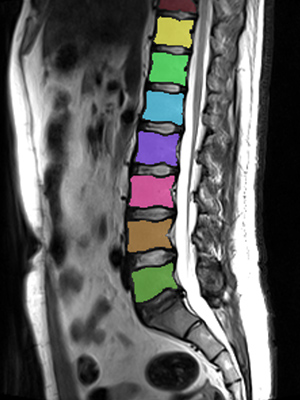}%
		\,%
		\includegraphics[width=0.242\textwidth]{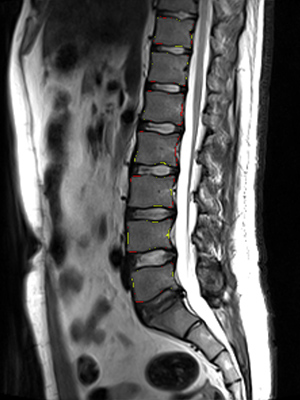}%
		\hfill%
		\includegraphics[width=0.242\textwidth]{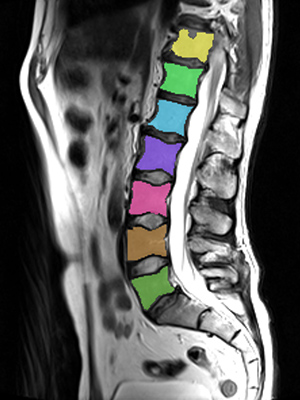}%
		\,%
		\includegraphics[width=0.242\textwidth]{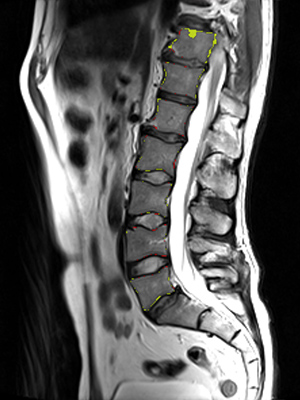}%
		\\[0.25em]
		(c) Lumbar spine MR
		\caption{Segmentation results in different types of images. The segmentations are shown both as color overlay with different colors for different instances (left), and as difference maps with oversegmentation errors marked in red and undersegmentation errors in yellow (right). Some images have been cropped to better show the vertebral column.}
		\label{fig:examples_differences}
	\end{figure*}

	Similar performance was achieved for vertebra segmentation in various CT datasets with an average Dice score of \SI{94.9(21)}{\percent} and for vertebral body segmentation in an MR dataset with an average Dice score of \SI{94.4(33)}{\percent}. Surface distances were lower on CT images compared to MR images (\SI{0.2(1)}{\milli\meter} vs.\ \SI{0.4(3)}{\milli\meter}), however, there were also fewer training scans available in the MR dataset. Figure~\ref*{fig:examples_differences} illustrates the magnitude of differences of the automatic segmentations from the ground truth segmentations.
	
\begin{figure}[t]
		\centering
		\includegraphics[width=8.75cm]{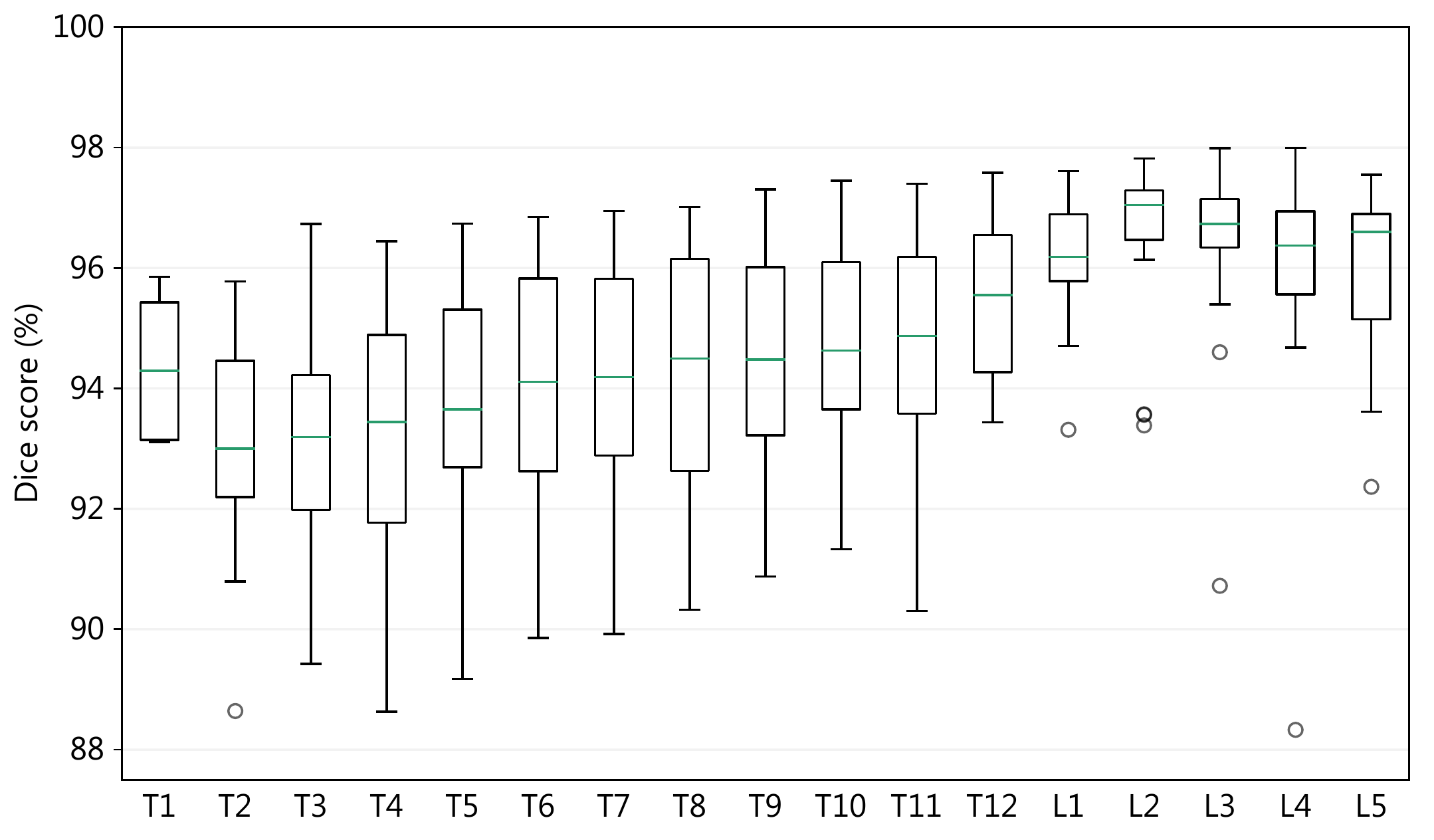}
		\\
		\includegraphics[width=8.75cm]{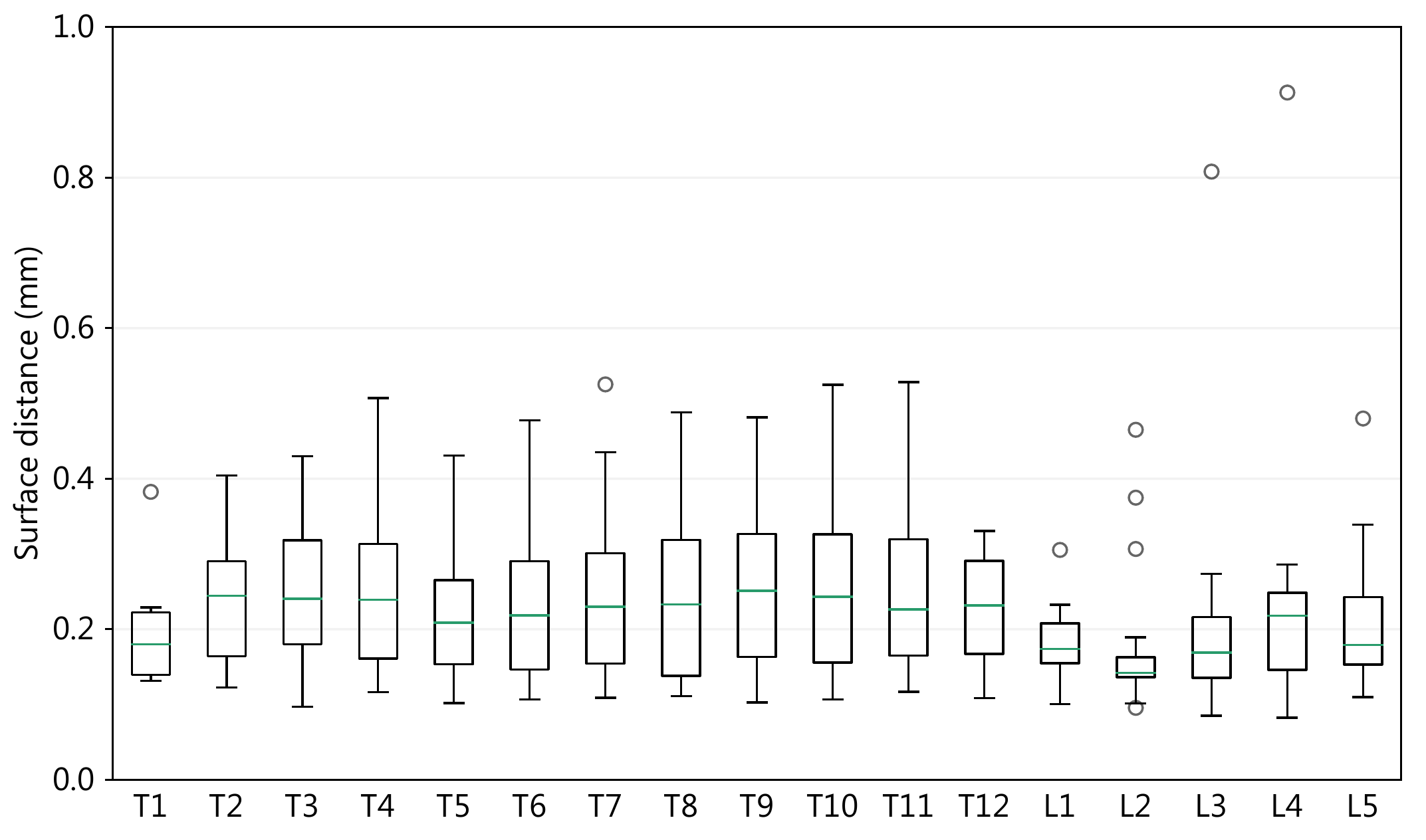}
		\caption{Box and whisker plots for per-vertebra Dice scores (top) and average absolute symmetric surface distances (bottom) in the CT evaluation set with 25 scans. Note that not all vertebrae were visible in every scan. Only completely visible vertebrae are included in the evaluation.}
		\label{fig:boxplots}
	\end{figure}

	In the CT datasets, the segmentation was more accurate on high-resolution dedicated spine scans of healthy subjects compared with low-dose low-resolution chest CT scans and scans of subjects with in some cases severe compression fractures. This is also visible in the segmentation performance stratified by vertebra (Figure~\ref*{fig:boxplots}). Segmentations were more accurate for the lumbar (L1-L5) than for the thoracic vertebrae (T1-T12), which are covered by the more challenging low-dose chest CT scans. Outliers among the lumbar vertebrae correspond to vertebrae from the xVertSeg.v1 dataset, which features a number of severely deformed lumbar vertebrae that are particularly challenging to segment.
	
	In comparison with other vertebra segmentation methods, our iterative instance segmentation approach outperformed previous methods on the thoracolumbar spine CT dataset as well as on the lumbar spine CT dataset. In both cases, there was a substantial improvement in average Dice score and especially also in the surface distance (Table~\ref*{tbl:results}). On the xVertSeg.v1 dataset with various fractured vertebrae, our method performed comparable to the method of \citet{Sekuboyina2017} and not as well as the method of \citet{Janssens2018}. However, both of these publications used a different separation between training and evaluation data and the results are therefore not directly comparable. For vertebral body segmentation in MR, our approach achieved on average higher Dice scores and lower surface distances than previous methods, but with higher variance compared to \citet{Korez2016}.
	Although the automatic segmentation was overall accurate on low-dose chest CT, the performance was still slightly below the level of interobserver variation (average difference of \SI{3.3}{\percent} in Dice score and \SI{0.1}{\milli\meter} in surface distance).
	
	The segmentation was optimized by minimizing a loss term based on false positive and false negative predictions. We additionally trained instances of the network using the categorical cross-entropy and the Dice coefficient as segmentation loss. However, the network optimized using the categorical cross-entropy converged in a state in which it predicted all voxels of any input patch as background. The network optimized using the Dice score performed substantially worse than the network trained with the proposed loss function, achieving for instance on low-dose chest CT scans an average Dice coefficient of \SI{89.9(68)}{\percent} and an average surface distance of \SI{0.5(7)}{\milli\meter} vs.\ an average Dice coefficient of \SI{93.1(15)}{\percent} and an average surface distance of \SI{0.3(1)}{\milli\meter} achieved by the proposed loss function.
	
	Furthermore, the proposed network architecture predicts from an image patch not only a segmentation mask, but additionally also the anatomical label and a score for the completeness of the vertebra. To investigate whether embedding these additional tasks into the same network influences the segmentation performance, we trained an instance of the network with identical training data and settings, but without the labeling and completeness classification paths. While we observed minor differences, the segmentation performance of the networks with and without additional tasks was overall comparable (Table~\ref*{tbl:results2}).
	
	\begin{table}[t]
		\caption{Comparison of the segmentation performance of a network consisting only of the segmentation path (S network) and a network consisting of the segmentation, the anatomical labeling and the completeness classification paths (S-L-C network), as proposed. Both networks traverse the spine upwards. Additionally, the segmentation performance is reported for the full network traversing downwards (S-L-C network~\downward).}
		\centering
		\footnotesize
		\begin{tabular}{lcc}
			\toprule
			& \multicolumn{2}{c}{\emph{Segmentation}} \\
			\addlinespace[0.2em]
			Dataset &
			Dice score (\si{\percent}) &
			ASSD (\si{\milli\meter}) \\
			\midrule
			Thoracolumbar spine CT &
			&   
			\\  
			\hspace{0.25cm} S network &
			\num{96.2(13)} &
			\surfacedistance{0.1(1)} \\
			\hspace{0.25cm} S-L-C network &
			\num{96.3(13)} &
			\surfacedistance{0.1(1)} \\
			\hspace{0.25cm} S-L-C network \downward &
			\num{96.2(14)} &
			\surfacedistance{0.2(1)} \\
			\addlinespace[0.2em]
			Lumbar spine CT\fns{1} & & \\
			\hspace{0.25cm} S network &
			\num{95.3(12)} &
			\surfacedistance{0.2(1)} \\
			\hspace{0.25cm} S-L-C network &
			\num{94.6(22)} &
			\surfacedistance{0.3(2)} \\
			\hspace{0.25cm} S-L-C network \downward &
			\num{95.1(11)} &
			\surfacedistance{0.2(1)} \\
			\addlinespace[0.2em]
			Low-dose chest CT & & \\
			\hspace{0.25cm} S network &
			\num{92.9(17)} &
			\surfacedistance{0.3(1)} \\
			\hspace{0.25cm} S-L-C network &
			\num{93.1(15)} &
			\surfacedistance{0.3(1)} \\
			\hspace{0.25cm} S-L-C network \downward &
			\num{91.0(82)} &
			\surfacedistance{0.6(11)} \\
			\addlinespace[0.2em]
			Lumbar spine CT & & \\
			\hspace{0.25cm} S network &
			\num{96.4(9)} &
			\surfacedistance{0.2(1)} \\
			\hspace{0.25cm} S-L-C network &
			\num{96.5(8)} &
			\surfacedistance{0.2(0)} \\
			\hspace{0.25cm} S-L-C network \downward &
			\num{96.3(9)} &
			\surfacedistance{0.2(1)} \\
			\bottomrule
			\addlinespace[0.2em]
			\multicolumn{3}{r}{%
				\scriptsize
				\fns{1}\,xVertSeg.v1 dataset
			} \\
		\end{tabular}
		\label{tbl:results2}
	\end{table}
	
	\subsection{Identification performance}
	
	The anatomical identification was correct for \SI{93}{\percent} (\nicefrac{206}{222}, $\kappa = 0.99$) of the vertebrae in the CT datasets. The labeling of one thoracolumbar spine CT scan was offset by one vertebra, resulting in incorrect labeling of all 16 visible vertebrae. The range T2-L5 was predicted, but this patient had only four lumbar vertebrae and the correct range was therefore T1-L4. In the MR dataset, anatomical identification succeeded in all cases ($\kappa = 1.0$). Overall, the labeling was correct in \SI{97}{\percent} of the scans.
	
	Even though a variety of scans with different sections of the spine were included in our evaluation, most still visualized anatomical landmarks such as the sacrum that potentially simplified the anatomical identification. Even low-dose chest CT scans have a fairly standardized field of view defined by the location and size of the lungs, which might have simplified vertebra identification. To evaluate the identification performance on arbitrary field of view images, we performed an experiment with randomly cropped images. For each of the 15 evaluation scans in the thoracolumbar spine CT and the chest CT dataset, we created two new images by randomly cropping the original image along the z-axis. These new images were minimally \SI{80}{\percent} and \SI{60}{\percent} smaller than the original image, but we ensured that they still contained multiple vertebrae by enforcing a minimum size of \SI{150}{\milli\meter} along the z-axis. Of the vertebrae visible in these cropped images,  \SI{93}{\percent} were correctly identified (\nicefrac{174}{187}; $\kappa = 0.98$). In the three images with mistakes, the labeling was offset by $\pm 1$, hence the high $\kappa$ score.
	
	\subsection{Completeness classification performance}
	
\begin{figure*}[t]
		\resizebox{\textwidth}{!}{
			\includegraphics{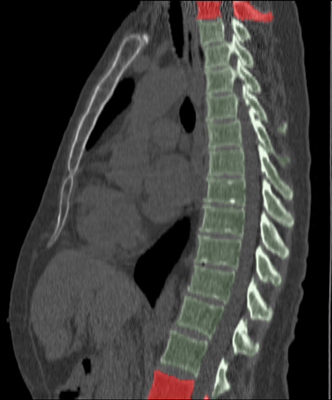}
			\quad
			\includegraphics{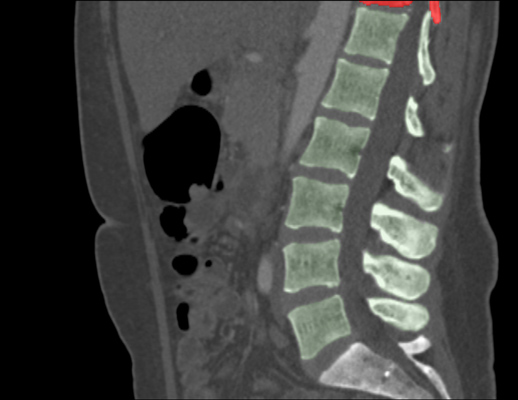}
			\quad
			\includegraphics{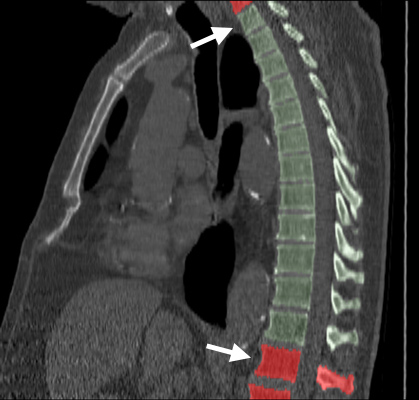}
		}
		\caption{Examples of vertebra completeness classification results. Vertebrae classified as completely visible are marked in light green and vertebrae classified as incompletely visible are marked in red.  Arrows indicate misclassified vertebrae. Shown are low-dose chest CT scans (left and right) and a lumbar spine CT scan (center).}
		\label{fig:examples_completeness}
	\end{figure*}

	None of the scans in any of the datasets visualized the entire spine, all scans contained vertebrae that were only partially visible due to the limited field of view. In the CT datasets, \SI{97}{\percent} of the vertebrae were correctly classified as completely or incompletely visible (\nicefrac{268}{275}).  In the MR dataset, all vertebrae were correctly classified. Most mistakes were made in low-dose chest CT scans, which was the type of scan with the least standardized field of view. Mistakes occurred only in the first or last visible vertebrae, near the boundary of the field of view, and often in vertebrae of which only a small part was missing from the scan (Figure~\ref*{fig:examples_completeness}). Notably, there were no mistakes where the network predicted an implausible sequence by classifying a vertebra between two completely visible vertebrae as incompletely visible, or vice versa.
	
	\subsection{Traversal direction}
	
	The networks in our experiments were trained to traverse upwards along the spine. To evaluate whether the direction of traversal influences the performance, we trained an additional instance of the network on the CT datasets to traverse downwards along the spine. We found no differences in anatomical identification and completeness classification of the detected vertebrae. However, while the segmentation performance was overall comparable with that of the same network traversing upwards, the segmentation performance deteriorated on the low-dose chest CT dataset (Table~\ref*{tbl:results2}). In multiple low-dose chest CT scans, the top-most vertebrae were less well segmented (Figure~\ref*{fig:examples_downwards_traversal}). When traversing downwards, these are the first vertebrae to be segmented. Notably, the traversal process was still set off correctly in all cases.
	
	In an additional experiment, we used two networks that were trained to traverse in opposite directions. The result of the upwards traversal was used to initialize the downwards traversal by starting the downwards traversal from the last vertebra detected during upwards traversal. We hoped that this would result in improved segmentation performance because it relieves the downwards network from detecting the first vertebra without any context information provided by the instance memory. However, this showed not to be beneficial and resulted in virtually identical segmentation performance compared to only upwards traversal.
	
	\subsection{Comparison with multiclass FCN}
	
\begin{figure*}[t]
    	\begin{minipage}[t]{0.49\textwidth}
    	    \centering%
    	    \begin{minipage}[t]{\columnwidth}
        		\footnotesize\centering%
        		\makebox[0.425\textwidth][c]{Upwards traversal}%
                \hspace{1ex}%
        		\makebox[0.425\textwidth][c]{Downwards traversal}%
        	\end{minipage}\\[1ex]%
    		\includegraphics[width=0.425\columnwidth]{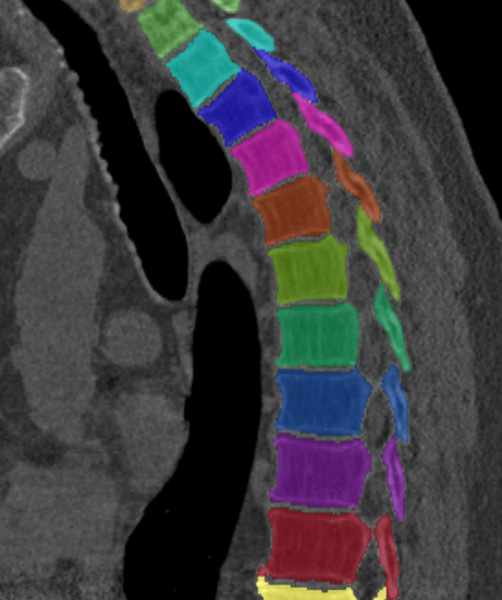}%
            \hspace{1ex}%
            \includegraphics[width=0.425\columnwidth]{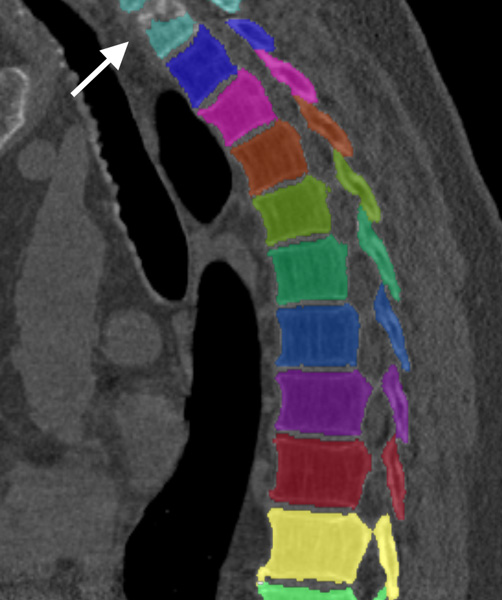}\\[1.5ex]%
    		\includegraphics[width=0.425\columnwidth]{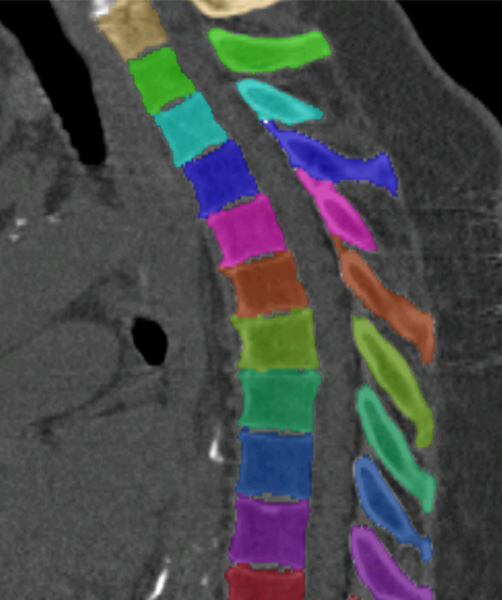}%
            \hspace{1ex}%
    		\includegraphics[width=0.425\columnwidth]{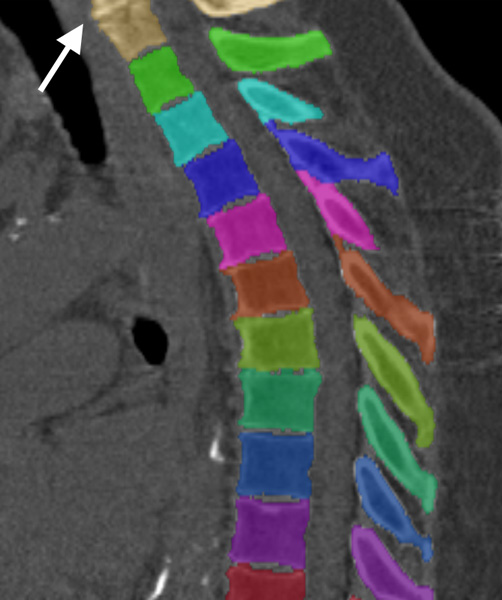}%
    		\caption{Segmentations obtained with the proposed iterative segmentation approach, comparing models trained for either upwards or downwards traversal. Both examples are low-dose chest CT scans (cropped). White arrows indicate segmentation errors in the top-most visible vertebrae, which occur more often when traversing downwards.}
    		\label{fig:examples_downwards_traversal}
    	\end{minipage}%
    	\hfill%
    	\begin{minipage}[t]{0.49\textwidth}
    	    \centering%
    	    \begin{minipage}[t]{\columnwidth}
        		\footnotesize\centering%
        		\makebox[0.425\textwidth][c]{Iterative FCN}%
        		\hspace{1ex}%
        		\makebox[0.425\textwidth][c]{Multiclass FCN}%
        	\end{minipage}\\[1ex]%
    		\includegraphics[width=0.425\columnwidth]{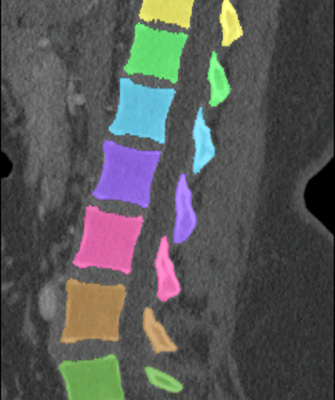}%
    		\hspace{1ex}%
    		\includegraphics[width=0.425\columnwidth]{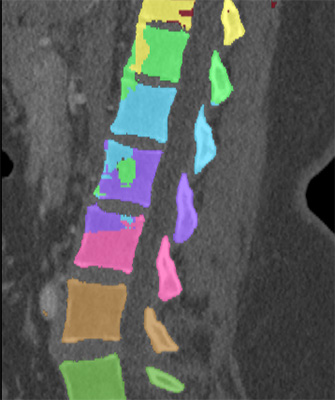}\\[1.5ex]%
    		\includegraphics[width=0.425\columnwidth]{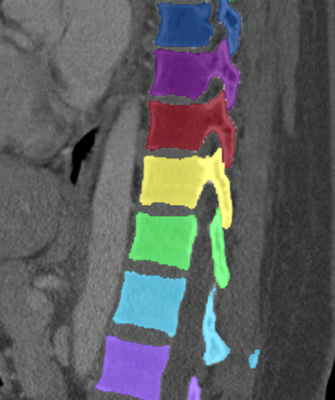}%
    		\hspace{1ex}%
    		\includegraphics[width=0.425\columnwidth]{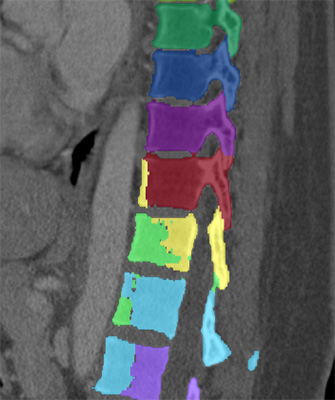}\\[1ex]%
    		\caption{Segmentations obtained with a multiclass FCN, thus without using the iterative segmentation strategy, compared with segmentations obtained with the proposed iterative approach. While the segmentations are overall fairly accurate, the individual vertebrae are not well separated.}
            \label{fig:examples_unet}
    	\end{minipage}
	\end{figure*}

    To verify that an instance segmentation approach is beneficial, we also trained the segmentation component of our network, which is a U-net like 3D FCN with skip connections (Figure~\ref*{fig:network}), to segment and identify the vertebrae using multiclass voxel classification instead of the proposed iterative binary segmentation. This network received an image patch of \num{128x128x128} voxels as input, but unlike in our iterative approach not a corresponding patch from the instance memory. The network had \num{25} output classes, corresponding to the \num{24} different vertebrae and a background class. At inference time, the patch was moved over the entire image in a sliding window fashion with overlapping windows so that multiple predictions were obtained for each voxel. Each voxel was eventually labeled with the class label that had the highest average probability. Using non-overlapping windows resulted in substantially worse performance. Overall, the multiclass FCN achieved an average Dice score of \SI{78.7(135)}{\percent} and an average ASSD of \SI{8.2(94)}{\milli\meter}. The anatomical identification was successful in \SI{89}{\percent} ($\kappa = 0.98$) of the vertebrae. While the segmentations were overall reasonably accurate with a Dice score of \SI{84.6(69)}{\percent} when the class labels were disregarded, the individual instances were often not well separated from each other (Figure~\ref*{fig:examples_unet}).
	
	\subsection{Runtime}
	
	The runtime of a single iteration step was about \SI{1}{\second} on standard hardware. The number of required iteration steps and thus the overall runtime per image depends on the size of the image, the number of visible vertebrae and their location within the image, which influences how many steps are initially needed to find the first vertebra. For instance, the low-dose chest CT scans covered on average \num{13.7} vertebrae and required \num{55} iteration steps. The thoracolumbar spine CT scans covered on average \num{17.8} vertebrae, but with a narrower field of view focused on the spine and required a comparable number of iteration steps (\num{56} on average). The average runtime per scan was about one minute in all datasets, excluding time required for loading the image and storing the results.
    
    \section{Discussion}
    
    This paper demonstrates that fully convolutional neural networks, which have been widely used for semantic segmentation \citep{Litjens2017}, are also capable of learning a complex instance segmentation task. Vertebra segmentation performed instance-by-instance required the network to learn to infer from an additional memory input which vertebra to segment and to ignore other vertebrae. Additionally, the same network was able to perform multiple tasks concurrently, namely vertebra segmentation, identification and classification to determine whether the vertebra was completely contained in the scan. This approach outperformed all methods that participated in the CSI 2014 spine segmentation challenge \citep{Yao2016} and performed better or comparable to state-of-the-art methods on other datasets. In a particularly challenging set of low-dose chest CT scans, the performance was close to the interobserver variability.
    
    The diverse selection of datasets that we used to evaluate the iterative segmentation approach demonstrates that this approach can cope with arbitrary fields of view, with low-resolution and low-dose scans, and with scans with normal as well as severely deformed vertebrae. The approach is entirely supervised, which enables transferring the same approach to other modalities and other segmentation tasks, which we demonstrated by applying the same approach without any modifications to vertebral body segmentation in T2-weighted MR images. Moreover, we demonstrated that the networks did not overfit to the datasets that were represented in the training data, but instead outperformed state-of-the-art methods on an entirely unseen datasets of lumbar spine CT scans.
    Previous methods for vertebra segmentation were often tailored to specific image types and evaluated on homogeneous datasets. Although our approach performed slightly worse than some of these methods on some of the datasets included in our evaluation, we demonstrated consistently high performance across multiple datasets.
    
    The proposed iterative vertebra-by-vertebra segmentation performed substantially better than a regular multiclass FCN similar to a 3D U-net \citep{Cicek2016}. However, the performance of the multiclass FCN exceeded our expectations and might further improve with more training data and with hardware that enables training of larger networks, or by including more context information via, e.g., a multi-scale approach \citep{Moeskops2016,Kamnitsas2017}. Even though the iterative approach requires the network to combine two inputs to identify a specific vertebra and ignore others, it also simplifies the segmentation problem from a multiclass into a binary voxel labeling task. The direct comparison of these two approaches indicates that individual instances of a target class of objects are better separated by a FCN if the network is trained to focus on individual instances. This strategy could also lead to improvements in other instance segmentation tasks, for instance, in histopathological image analysis.
    
    We combined multiple tasks into a single network, which helps to simplify both training and inference: only a single network needs to be trained, and at inference time, each patch needs to be passed only through one network to obtain multiple predictions, one for each distinct task. Even though the segmentation path and the identification and completeness classification paths shared part of the network, we did not find that this improved the segmentation performance. However, the segmentation performance did also not deteriorate when these additional tasks were added, which indicates that the proposed combination into a single network is useful. The proposed network architecture with the additional output paths uses more GPU memory than task specific networks would and therefore more strongly limits the maximum number of filter per layer and the depth of the network. Future hardware generations with larger memory will enable training of larger networks, which might lead to further performance improvements.
    
    The proposed spine traversal strategy can be applied to segment the vertebrae from top to bottom or vice versa. However, we observed better performance for upwards compared to downwards traversal on low-dose chest CT scans with differences mostly in the upper vertebrae. The size of the vertebrae increases from cervical to lumbar vertebrae, i.e., from top to bottom. Hence, when traversing downwards, the first vertebra that needs to be found, without additional information that could be derived from the memory input, is the smallest vertebra that is visible in the scan. Additionally, the region around the uppermost visible vertebrae is often affected by low-dose artifacts in low-dose chest CT scans, which makes starting the traversal from these vertebrae especially challenging. Because there were no substantial differences between downwards and upwards traversal in the other CT datasets, upwards traversal presents overall the more robust strategy.
    
    Even though we used a variety of datasets in our experiments, some important types of data were not present. These include scans covering the cervical vertebrae and scans of patients with implants near the spine, such as pedicle screws. However, since our approach is entirely supervised and trained end-to-end, it will likely be able to handle these cases if sufficient training examples are available. Furthermore, there were no scans of patients with irregular numbers of vertebrae present in the training set, which caused mislabeling of the vertebrae in one patient with irregular number of vertebrae in the evaluation set. Ensuring accurate anatomical labeling for such cases could be an interesting direction for further research. However, the high kappa scores indicate that the labeling would be minimally offset in such cases. Depending on the exact clinical application, the impact of such labeling mistakes would therefore be limited.
    
    Segmentation of the vertebrae one after the other, using information about the already segmented vertebrae as a prior, is inherently susceptible to cascading failure. Failure to find or correctly segment a single vertebra may cause failure to find or correctly segment all subsequent vertebrae. There is additionally no element that explicitly ensures that the predicted segmentation mask covers only a single vertebra if multiple are visible. While we did not observe these kind of failures in our evaluation, they are likely to occur in images with extreme anatomical abnormalities or severe imaging artifacts. Refinement of the labeling through a maximum likelihood approach suffers from a similar weakness: Except for cases with irregular number of vertebrae, the labeling can only be entirely correct or entirely offset, even if the correct labels were predicted for some of the vertebrae. This limitation could potentially be addressed in the future by employing a more sophisticated global labeling model, e.g., based on Markov models.
    
    Manual vertebra segmentation is a time-consuming and tedious task, requiring annotation times of about 40 to 60 hours per scan in low-dose chest CT and even longer in scans with higher resolution and coverage of more vertebrae. Semi-automatic segmentation proved to be an effective strategy for generating reference segmentations for network training from only few manual reference segmentations. Especially if little training data is available, additional priors or model fitting steps could help stabilize the performance. These could be statistical knowledge about typical sizes or shapes of vertebra, or additional fitting of a deformable surface mesh model to the segmentation results \citep{Korez2016}.
    
    Precise segmentation and identification of the vertebrae from CT and MR scans enables automatic spine analysis, notably also in images that were originally not intended for spine imaging. For instance, our iterative approach achieved a segmentation and identification performance on low-dose chest CT scans that is likely sufficient to analyze the shape of the vertebral bodies for detection of compression fractures. This could enable opportunistic screening for early signs of osteoporosis in lung cancer screening programs, in addition to screening for pulmonary abnormalities.
    
    In conclusion, this paper presents an iterative instance-by-instance approach to vertebra segmentation and anatomical identification. This approach is fast, flexible and accurate across a large variety of both dedicated as well as non-dedicated spine scans.
        
    \section*{Acknowledgements}
    
    \noindent   
    We would like to thank the organizers of the CSI 2014 spine segmentation challenge, the Laboratory of Imaging Technologies at the University of Ljubljana and the authors of the MR dataset for making scans and reference segmentations publicly available. We are furthermore grateful to the United States National Cancer Institute (NCI) for providing access to NCI’s data collected by the National Lung Screening Trial. The statements contained in this publication are solely ours and do not represent or imply concurrence or endorsement by NCI.

\end{document}